\documentclass[11pt]{article}
\usepackage[a4paper,margin=2.5cm]{geometry}
\usepackage{amsmath,amssymb}
\usepackage{booktabs}
\usepackage{graphicx}
\usepackage[numbers,sort&compress]{natbib}
\usepackage{xurl}
\usepackage[hidelinks]{hyperref}
\input{tables/numbers.tex}
\title{How Far Can the Gemma~4 Speech Encoder Go Without Changing Its Weights?\\ Saudi and Gulf Arabic Speech Recognition with a Frozen and an Adapter-Augmented Encoder}
\author{Amr Abdelkhalik\\ Independent Researcher, Egypt\\ \texttt{amr.khalil.abdelkhalik@gmail.com}}
\date{}
\begin{document}
\maketitle
\begin{abstract}
We study the 305M-parameter audio encoder shared by Gemma~4 and EmbeddingGemma~2 (bit-identical to that of Gemma~4 E2B except for the final audio-embedding projection) for Saudi and Gulf Arabic speech
recognition, without changing any of its pretrained weights; the adapters of our adapted system are additive and kept separate. A frozen system trains only a 22.6M-parameter CTC reader over a learned mix of four intermediate encoder
layers and the encoder output; an adapted system adds 3.1M low-rank adapter parameters, a 41.5M-parameter hybrid CTC/attention reader and joint decoding
with a sub-word language model. Against Cohere Transcribe Arabic (2.07B parameters), re-run on identical audio, the frozen system
trails in word error rate (WER) but is not significantly different in character error rate (CER) on five of six
human-labelled conditions, while running \R{speed.ratio.cpu} times faster on CPU with greedy decoding. The adapted system is not significantly different in WER on a public Saudi customer-service call
(\R{ad.nexdata_cs_sample.clean16k.jointlm.wer}\% against \R{res.cohere.nexdata_cs_sample.clean16k.wer}\%) and on clean Emirati speech,
is significantly better on telephone-coded Emirati speech, has a significantly lower CER on all six conditions of these sets, and beats a
0.6B recogniser fine-tuned on SADA in WER on all six; it still trails Cohere in WER on long-form Saudi speech. On a blind,
human-corrected public call-centre sample it ranks \R{api.rank_ours} of \R{api.n_systems} systems, behind three commercial APIs (significantly only one)
and ahead of the other self-hostable systems by point estimate. An optimised joint decoder runs it at \R{fast.p4.rtfx} times real time on one GPU. We
also find, on our data, that references made by correcting machine drafts favour the drafting system. Code and public-set results are released.

\end{abstract}
\section{Introduction}
\label{sec:intro}

Arabic speech recognition is hardest where it matters most commercially: dialectal speech over the telephone. Saudi and Gulf call centres mix regional dialects, code-switching, and spoken digit strings such as identity and account numbers. All of this arrives on 8 kHz narrow-band audio. Public benchmarks confirm the difficulty. On the Saudi SADA test set, the best open systems in the Open Universal Arabic ASR Leaderboard still reach word error rates (WER) above 40\% \cite{arabic_leaderboard}. A dedicated benchmark of Arabic call-domain speech reports WERs near 50\%, and this holds even for large commercial recognisers \cite{maqsam_call}.

The usual remedies are to fine-tune a large pretrained recogniser \cite{whisper,mms,wav2vec2,owsm} or to train a dedicated Arabic model on many hours of labelled speech \cite{weak_arabic_asr,artst}. Both demand substantial compute for every new domain. A cheaper option keeps a strong speech encoder frozen and trains only a small task head on top of it. Probing benchmarks such as SUPERB and ML-SUPERB \cite{superb,ml_superb} show that frozen self-supervised encoders retain enough phonetic information for recognition. Layer-wise analyses go further and reveal that this information peaks in intermediate layers rather than at the output \cite{pasad_layerwise,pasad_words}.

Open multimodal language models now ship with their own speech encoders. Gemma~4 \cite{gemma4} attaches a 305M-parameter Conformer audio encoder, derived from the Universal Speech Model design \cite{usm}, to a text decoder, and the same encoder has since been reused elsewhere: Google states that EmbeddingGemma~2, a multimodal embedding model, ``shares its text tokenizer and audio encoder'' with Gemma~4 \cite{eg2_blog,embeddinggemma2}. We confirmed this directly for Gemma~4 E2B. Of the 752 audio-tower and audio-embedding tensors in the released EmbeddingGemma~2 checkpoint, 751 are bit-identical to those of Gemma~4 E2B \cite{gemma4_e2b}; the only difference is the audio-embedding projection into each model's text embedding space, which lies outside the audio tower and which a recogniser does not use. The Gemma~4 report describes the same 305M-parameter encoder for the larger E4B model \cite{gemma4}; we did not compare E4B's weights. The encoder is therefore a component that developers may already be running, inside an assistant model or an embedding service, and it is released under the Apache-2.0 licence. Whether it can be reused for Arabic transcription without fine-tuning has, to our knowledge, not been studied.

This paper asks a narrow question. How far can this speech encoder go on Saudi and Gulf Arabic, including telephone speech, when its weights stay unchanged or are adapted only through a small number of added parameters? We make five contributions.
\begin{enumerate}
\item A \emph{frozen} recipe: the unchanged Gemma~4 audio encoder (as distributed with EmbeddingGemma~2), a learned mix over four of its hidden layers and its output, a 22.6M-parameter transformer reader trained with connectionist temporal classification (CTC) \cite{ctc}, and a word 4-gram language model in beam search \cite{kenlm,pyctcdecode}.
\item An \emph{adapted} recipe that keeps the encoder's weights frozen but adds 3.1M low-rank adapter parameters \cite{lora} to six middle layers, with a 41.5M-parameter hybrid CTC/attention reader \cite{hybrid_ctc_att}, self-conditioned CTC \cite{selfcond} and joint CTC/attention decoding \cite{joint_ctc_att_decoding} with a sub-word language model.
\item An evaluation on SADA, three further human-labelled Saudi and Gulf test sets, a blind human-corrected public call-centre sample with fifteen recognisers including eleven commercial services, and real call-centre recordings, against open recognisers re-run on identical audio: Cohere Transcribe Arabic (2.07B parameters) \cite{cohere_transcribe} and Nemotron~3.5 ASR (0.6B) \cite{nemotron_asr}, zero-shot and fine-tuned.
\item We present a faster implementation of joint CTC/attention decoding with a language model. On one GPU, it raises throughput from \R{ref.joint_orig} to \R{fast.p4.rtfx} times real time. Accuracy is essentially unchanged. We also measure speed and memory on GPU and CPU under a single protocol. Paired bootstrap intervals \cite{bisani_bootstrap} and N-best oracle analyses then separate model errors from ranking errors.
\item Call-centre references produced by correcting machine drafts are mostly the unedited drafts. Such references favour the system that produced them, so we report results split by reference origin.
\end{enumerate}

The results depend on the measure. The frozen system stays behind the 2B-parameter recogniser in WER on every test set. Its character error rate, however, is not significantly different from that recogniser's on five of the six human-labelled conditions, and it runs several times faster on CPU. The adapted system is not significantly different from the 2B recogniser in WER on the public customer-service call and on clean Emirati speech, and better on telephone-coded Emirati speech. On the further human-labelled sets it is significantly lower in CER across all six conditions. In WER, however, it remains behind on long-form SaudiTalk speech. Section~\ref{sec:related} reviews related work, Section~\ref{sec:method} describes the systems, Section~\ref{sec:setup} the data and protocol, Section~\ref{sec:results} the results, Section~\ref{sec:discussion} discusses limitations, and Section~\ref{sec:conclusion} concludes.

\section{Related Work}
\label{sec:related}

\subsection{Dialectal and Telephone Arabic ASR}
Arabic recognition research has moved from Modern Standard Arabic broadcast speech towards dialects. SADA \cite{sada_corpus} provides several hundred hours of Saudi television speech with human transcripts, and it has become the standard Saudi test bed. Fine-tuning results on its clean test subset are reported in \cite{sada_asr}. Casablanca \cite{casablanca} covers eight Arabic varieties, including Emirati and Yemeni, with human transcripts and code-switching annotations. The Open Universal Arabic ASR Leaderboard \cite{arabic_leaderboard} evaluates open models on several of these sets under a common text normalisation. Large-scale weakly supervised training on Arabic is studied in \cite{weak_arabic_asr}, while ArTST \cite{artst} pre-trains a joint Arabic speech--text model. For the call domain, \cite{maqsam_call} released a benchmark of Arabic call-centre style speech, yet its audio is 16 kHz. No public Saudi corpus of real 8 kHz call-centre speech of useful size exists, only small free samples, so we combine real call recordings, which we cannot release, with public human-labelled sets and with telephone-coded versions of them.

\subsection{Frozen Encoders and Layer Analysis}
Self-supervised encoders such as wav2vec 2.0 \cite{wav2vec2} and w2v-BERT \cite{w2vbert,seamless} are commonly fine-tuned end to end. SUPERB \cite{superb} and ML-SUPERB \cite{ml_superb} instead evaluate them frozen, feeding a learned weighted sum over layers into a light downstream model. This protocol is the basis of our reader. Frozen encoders of supervised recognisers are also strong general features. Whisper-AT \cite{whisper_at}, for example, tags audio events from the frozen Whisper encoder. Pasad et al. \cite{pasad_layerwise,pasad_words} showed that phonetic and word information is strongest in middle layers and weaker at the top of pretrained speech models.

\subsection{Speech Encoders in Multimodal and Embedding Models}
Recent multimodal language models attach a speech encoder to a text decoder. SLAM-ASR \cite{slam_asr} showed that a frozen encoder, a small projector and a language model can give strong English recognition. The Gemma 4 report \cite{gemma4} describes an audio encoder of about 305M parameters. That encoder uses two down-sampling convolutions and 12 Conformer layers \cite{conformer}, producing one frame every 40 ms. A separate line of work trains speech encoders for cross-modal or semantic alignment, including SAMU-XLSR \cite{samu_xlsr}, SONAR \cite{sonar}, SpeechCLIP \cite{speechclip} and CLAP \cite{clap}. Those models are optimised for utterance-level representations rather than transcription. The encoder studied here belongs to the first group, because it is the encoder of an assistant model reused unchanged in an embedding model. We are not aware of a published Arabic recognition study of it.

\subsection{Reader Architectures and Decoding}
CTC \cite{ctc} gives a non-autoregressive output layer. Hybrid CTC/attention models \cite{hybrid_ctc_att}, intermediate CTC losses \cite{interctc} and self-conditioned CTC \cite{selfcond} relax the conditional independence of CTC. Joint CTC/attention decoding \cite{joint_ctc_att_decoding} scores attention hypotheses with CTC prefix probabilities. Shallow fusion adds an external language model to sequence-to-sequence search \cite{shallow_fusion}. External n-gram models remain a cheap way to add word-level context to CTC output \cite{kenlm}. Our frozen system keeps the reader deliberately simple, so that the measurement isolates the encoder. The adapted system then combines these established components.

\section{Method}
\label{sec:method}
\begin{figure*}[t]\centering\includegraphics[width=\textwidth]{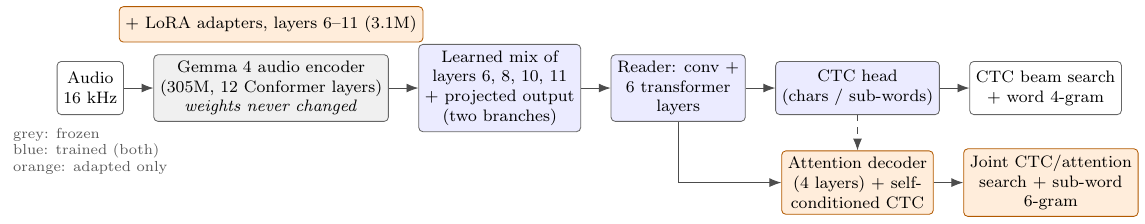}
\caption{The two systems. The Gemma~4 audio encoder is never changed (grey). Both systems train a layer mix, a reader and a CTC head (blue); the adapted system adds low-rank adapters inside the encoder, an attention decoder with self-conditioned CTC and joint CTC/attention search (orange).}\label{fig:arch}\end{figure*}

Figure~\ref{fig:arch} gives an overview of both systems.

\subsection{Frozen Encoder}
\label{sec:frozenenc}
The model's own feature extractor converts audio at 16 kHz into log-mel features. The audio encoder, loaded from the EmbeddingGemma 2 checkpoint, runs in bfloat16 with all weights frozen. It has 12 Conformer-style layers of width 1024, followed by an output projection to 1536 dimensions, and it emits 25 frames per second. Its released configuration uses chunk-wise local self-attention over chunks of 12 frames with no right context, so no frame attends beyond its current 480 ms chunk: the attention is block-causal with up to 480 ms of look-ahead. The reader described next is not causal. The reader takes the hidden states of layers 6, 8, 10 and 11 and the projected 1536-dimensional output. It also takes a sixth input, labelled ``layer 12'' in our code; because of how the library returns hidden states, this input is in fact the projected output, so that representation is fed twice, through two separate normalisation and projection branches, and the raw output of layer 12 is never used. This corrects the description in the first version of this paper; it does not change any result. A preliminary probe guided this subset; in it, the input labelled layer 12, that is the projected output, scored best. The other layers were not tried in the mix. Telephone audio is up-sampled to 16 kHz before feature extraction.

\subsection{Reader}
Each selected representation $h_\ell$ is normalised and projected to 512 dimensions. The projections are then combined with
softmax-normalised weights:
\begin{equation}
x_t = \sum_{\ell} \frac{\exp(a_\ell)}{\sum_{k}\exp(a_k)}\,W_\ell\,\mathrm{LN}(h_{\ell,t}),
\label{eq:mix}
\end{equation}
where the $a_\ell$ are learned scalars initialised at zero. The mixed sequence passes through a residual depth-wise convolution
(kernel 5), six pre-norm bidirectional transformer layers (512 dimensions, 8 heads, feed-forward size 2048, dropout 0.1) and a linear CTC output
layer over 68 symbols: the Arabic letters left after normalisation, digits, the word boundary and the CTC blank. The reader has 22.6M
parameters. During training, two random time spans per utterance, each up to 5\% of its length, are zeroed in the mixed features, in
the spirit of SpecAugment \cite{specaugment}.

\subsection{Training}
We train the reader with the CTC loss. We use AdamW with a learning rate of $3\times10^{-4}$ and weight decay 0.01, under a one-cycle schedule. Batches hold up to 6,000 encoder frames. Encoder features are computed on the fly, so the telephone simulation below can be applied afresh in each epoch. We select the model by validation character error rate on held-out calls.

\subsection{Telephone Simulation}
To expose the reader to narrow-band speech, wideband training audio is degraded on the fly. The audio is first down-sampled to 8 kHz, then encoded and decoded with G.711 $\mu$-law \cite{g711}, and finally up-sampled back to 16 kHz. The telephone versions of the test sets are produced by this same deterministic procedure.

\subsection{Decoding}
We report greedy CTC decoding and beam search with a word 4-gram language model \cite{kenlm} through pyctcdecode \cite{pyctcdecode} (beam 64). The language model is trained on the training-call transcripts and the SADA training transcripts, after the same normalisation as the scoring. Its weight and word-insertion bonus are tuned on the validation calls only, and the selected values are 0.5 and 2.0. Test data are never used for this tuning.

\subsection{Adapted System}
\label{sec:adapted}
The adapted system keeps every encoder weight frozen but inserts low-rank adapters \cite{lora} (rank 16, scaling 32/16). They are placed first in the query, key, value and output projections of encoder layers 6 to 10 and later also in the feed-forward and convolution modules of layers 6 to 11, adding 3.1M trainable parameters in total. Its output units are 1,024 SentencePiece sub-words \cite{sentencepiece} instead of characters. The reader keeps the layer mix, convolution and six transformer layers of the frozen system and adds a CTC head and a four-layer autoregressive transformer decoder (512 dimensions, 8 heads) with gated sinusoidal positions; the reader has 41.5M parameters. Training proceeds in five warm-started stages: (1) sub-word CTC with the attention adapters; (2) the hybrid reader, initialised from stage 1, trained with CTC and attention losses for 6 epochs; (3) the decoder alone on the frozen reader, with label smoothing of 0.05 on content tokens, none on the end-of-sequence token and a weight of 1.5 on it, so that the decoder learns to stop; (4) all reader and adapter parameters jointly for 3 epochs with loss weights 0.3 (CTC), 0.5 (attention) and 0.2 (self-conditioned CTC \cite{selfcond}, a shared head after reader layers 2 and 4 whose posteriors are fed back through a zero-initialised projection); and (5) the same joint training for 4 further epochs after adding zero-initialised adapters to the feed-forward and convolution modules, with 80 h of SADA per epoch. Every warm-started run first re-scores the validation set and must reproduce the source model's CTC result within one point. Each machine-labelled call window is weighted by the agreement of its two labels: windows whose labels differ by at most 25\% CER get weight 1, by at most 35\% weight 0.5, by at most 45\% weight 0.25, and the rest are dropped; SADA utterances marked as unclear are removed.

The adapted system is decoded in two ways. The first, CTC beam search, uses a word 4-gram (weight \R{ad.ctc_alpha}, word bonus \R{ad.ctc_beta}). The second, joint CTC/attention beam search \cite{joint_ctc_att_decoding} (CTC weight 0.4), adds a 6-gram over the model's own sub-words by shallow fusion \cite{shallow_fusion} (weight \R{ad.lmw}, length bonus \R{ad.lenbonus}). Both language models are trained on the training-call transcripts, SADA training transcripts and MASC training transcripts \cite{masc}, with all validation and test calls excluded. All decoding weights, and the choice between two language-model variants, were selected on the validation calls only. That selection followed stage 4 with a beam of 10, where the chosen variant scored WER \R{ad.val.lm1}\% on validation against \R{ad.val.lm2}\% for the alternative. The choices were then kept unchanged for stage 5. Test results use a beam of 16 and its top hypothesis.

\section{Experimental Setup}
\label{sec:setup}

\subsection{Training Data}
\paragraph{Call-centre speech} The in-domain data are real Saudi call-centre recordings, taken as single-party legs of 8~kHz
telephone calls and stored at 16~kHz. We cut them into windows of up to 30~s. Training labels are machine-generated: each window
carries one transcript from Cohere Transcribe Arabic and one from a commercial cloud recogniser from Deepgram, the same vendor as the Deepgram Nova-3 service benchmarked in Table~\ref{tab:apis}, and in each epoch one of
the two is sampled per window. Calls reserved for validation and testing are kept out of both training and the language-model text.
After filtering, 6{,}050 training windows with 12{,}089 labels remain. The validation set holds 295 windows drawn from 147 calls, and because these validation labels are also machine transcripts,
validation scores serve only to select models.

\paragraph{Voicebot calls} The adapted system additionally uses 1.63 h of real Saudi voicebot telephone calls (67 calls, 2,011 voice-activity segments), repeated four times per epoch. Their labels are machine transcripts aligned to Cohere segment hypotheses. Segments whose machine transcript and Cohere hypothesis differ by more than 60\% WER are dropped.

\paragraph{SADA} The SADA training set \cite{sada_corpus} supplies human-labelled Saudi speech: \R{sada.n.train} utterances (\R{sada.h.train} h), of which 60 h are sampled in each epoch. In the \emph{telephone} variant, each SADA utterance is degraded as described above, and in the \emph{clean} variant it is used as is.

\subsection{Systems}
We train three readers. The \emph{calls-only} reader is trained on the call-centre windows for 20 epochs, and the best validation epoch falls at 9. Both the \emph{telephone-SADA} reader and the \emph{clean-SADA} reader begin from the calls-only reader and continue for 6 further epochs. Their training data consist of the calls plus SADA, and SADA is degraded to telephone quality in one case but kept clean in the other. For both, the best validation epoch is 3. We trained each configuration once. The adapted system of Section~\ref{sec:adapted} starts from the same data, with the additions described above.

Two open recognisers serve as baselines. Both were run on exactly the same audio. They were scored with the same normalisation.
\begin{itemize}
\item \textbf{Cohere Transcribe Arabic} (07-2026, 2.07B parameters, Apache-2.0) \cite{cohere_transcribe,cohere_blog}, a dedicated
Arabic encoder--decoder recogniser. We use greedy decoding on voice-activity segments of at most 25~s, with a repetition penalty of
1.2 and no repeated 4-grams to prevent the looping that the model otherwise shows on silence.
\item \textbf{Nemotron~3.5 ASR streaming 0.6B} \cite{nemotron_asr}, a cache-aware streaming FastConformer transducer. We use it zero-shot and after full fine-tuning on SADA only, in two stages of 4,000 steps each (learning rate $10^{-5}$, cosine schedule): first on clean SADA, then on about 50 h of SADA coded with GSM to simulate telephony. It never saw the call-centre data. Decoding uses beam search of width 8, with the attention-context setting of 56 frames of left and 13 of right context.
\end{itemize}
Cohere Transcribe Arabic supplied one of the two machine labels for the call-centre training data, so the frozen-encoder system is partly distilled from it. The fine-tuned Nemotron model was trained on SADA alone.

\subsection{Test Sets}
\begin{itemize}
\item \textbf{SADA test} set: \R{sada.n.test} human-labelled utterances of Saudi television speech \cite{sada_corpus} (after excluding one Latin-script reference), in clean and telephone-coded form.
\item The \textbf{Saudi customer-service sample} is the free portion of a commercial corpus \cite{nexdata_sample}. It consists of one 13-minute call leg, cut at its human time stamps into \R{res.n.nexdata_cs_sample} turns. It is the closest public match to our domain. Still, it comes from a single call and a single speaker.
\item \textbf{Casablanca, Emirati}: the \R{res.n.casablanca_uae} human-labelled utterances of the Emirati test split \cite{casablanca}.
\item \textbf{SaudiTalk} comprises \R{res.n.sauditalk_longform} human-verified recordings in Hijazi, Ha'il and Southern dialects \cite{sauditalk}, amounting to about 6 h in total. Transcripts cover whole clips and carry no time stamps, so all systems decode voice-activity segments of at most 25 s, and the segment outputs are then joined per clip.
\item The blind call-centre sample contains \R{do.nclips} clips, totalling \R{do.minutes} minutes and \R{do.nwords} reference words. The clips come from two one-sided mobile recordings in a public Saudi call-centre dialogue sample \cite{dataocean_saudi}. Clips were cut at voice-activity boundaries. Each clip received a draft transcript from one of three systems, assigned at random in equal shares and hidden from the annotator. The author, a native speaker, then corrected each draft into a verbatim reference. Each system is scored only on clips its own draft did not produce. Pairs of systems are compared only on clips drafted by neither of them.
\item The call-centre set contains \R{gold.n} windows drawn from \R{gold.calls} held-out calls in the in-house recordings. The windows were deliberately selected to over-represent spoken numbers (\R{gold.n.numbers} windows) and windows on which the two machine transcripts disagreed most. Consequently, they are harder than typical call audio.
\end{itemize}
Every test set except the in-house calls is also scored after telephone coding.

\paragraph{How the call-centre references were made} An annotator listened to each window in a correction tool prefilled with the longer of the two machine transcripts, and was asked to make it verbatim. In \R{gold.src.n.unchanged} of the \R{gold.n} windows the draft was accepted unchanged: \R{gold.src.n.coh} Cohere drafts and \R{gold.src.n.dg} drafts from the commercial cloud recogniser. Only \R{gold.src.n.edited} were edited. The references are therefore mostly machine text checked by a single listener, and they favour the drafting systems. Cohere's own draft scores \R{gold.firstpass_cohere.wer}\% WER against them, whereas the same model re-run on identical audio scores \R{gold.cohere.wer}\%. The two Cohere runs also differ in decoding settings, so not all of this gap is due to drafting; the split by reference origin below is the cleaner evidence. We treat this set as secondary and report it split by reference origin.

\subsection{Metrics and Statistics}
Text is normalised identically for every system: Unicode NFKC, removal of diacritics and tatweel, unification of alef, alef maqsura,
taa marbuta and hamza carriers, collapse of letters repeated three or more times, removal of Arabic punctuation (comma, semicolon, question mark and similar marks), and removal of all characters other than Arabic
letters, digits and spaces. Digits and spelled-out numbers are not converted into each other, so a number written in words by one system and in digits by the reference counts as an error. Our main WER and character error rate (CER, with spaces removed) are computed per utterance and averaged over utterances, which weights every utterance equally; we also report corpus-level WER (total errors over total reference words). CER ignores word boundaries, while WER counts a wrongly split or merged word as an error. Differences between
systems are tested with a paired bootstrap over utterances (10{,}000 resamples) \cite{bisani_bootstrap}; on the call-centre set we also resample whole calls, since several windows can come from one call. We report the bootstrap mean difference
and its 95\% percentile interval; these means can differ from the difference of the two printed scores in the last digit. Each interval is an individual 95\% interval, uncorrected for multiple comparisons, so an interval that only just excludes zero is weak evidence on its own. References written mainly in Latin script (Arabic letters fewer than twice the Latin letters), which are translations rather than transcripts, are excluded; this removes one SADA utterance and none elsewhere. Validation figures quoted for model development were computed during development with an earlier normaliser that kept Arabic punctuation; every test figure in this paper uses the final normaliser, applied to saved hypotheses of every system.

\section{Results}
\label{sec:results}

\subsection{SADA}
Table~\ref{tab:sada} reports the SADA test set. Adding SADA to training lowers WER with the language model from \R{sada.v2.sada_test_clean16k.lm.wer}\% to \R{sada.v3_tel.sada_test_clean16k.lm.wer}\% on clean audio, and from \R{sada.v2.sada_test_tel8k.lm.wer}\% to \R{sada.v3_tel.sada_test_tel8k.lm.wer}\% on telephone audio. The two SADA variants differ by less than one WER point. Training on telephone-coded SADA is \R{sada.tel_vs_wide.tel} points better on telephone audio, while training on clean SADA is \R{sada.tel_vs_wide.clean} points better on clean audio. In CER, telephone training helps more on telephone audio (\R{sada.v3_tel.sada_test_tel8k.lm.cer}\% against \R{sada.v3_wide.sada_test_tel8k.lm.cer}\%). The simulated coding is therefore useful mainly at the character level.

\begin{table*}[t]\centering\small\caption{SADA test set (4,971 utterances): WER and CER (\%) on the original 16~kHz audio and after simulated telephone coding (8~kHz).}\label{tab:sada}
\resizebox{\textwidth}{!}{\begin{tabular}{lrrrrrr}\toprule
& \multicolumn{3}{c}{Clean 16 kHz} & \multicolumn{3}{c}{Telephone 8 kHz}\\
& \multicolumn{2}{c}{WER} & CER & \multicolumn{2}{c}{WER} & CER \\
Reader & greedy & +LM & +LM & greedy & +LM & +LM \\\midrule
Calls only & 71.06 & 59.84 & 32.44 & 74.16 & 63.55 & 36.12 \\
+ SADA, telephone-coded & 59.98 & 48.55 & 24.16 & 63.33 & 51.58 & 26.85 \\
+ SADA, clean & 60.06 & 48.36 & 24.41 & 63.45 & 52.25 & 28.15 \\
\bottomrule\end{tabular}}\end{table*}

Table~\ref{tab:sadacmp} compares all systems on SADA test under the same normaliser. For context only, the leaderboard \cite{arabic_leaderboard} reports \R{lit.nemo_lm}\% WER on SADA test for an NVIDIA Conformer-CTC model with a language model, and \R{lit.whisper}\% for Whisper large-v3 \cite{whisper}. Cohere reports \R{lit.cohere}\% for its model \cite{cohere_blog}. These figures use a different normalisation and corpus-level aggregation, so they are not directly comparable.

\begin{table}[t]\centering\small\caption{SADA test (4,971 utterances), WER / CER (\%) of every system under the same normaliser. Our systems use CTC search with the word 4-gram.}\label{tab:sadacmp}
\begin{tabular}{lcc}\toprule
System & Clean & Telephone \\\midrule
Frozen (telephone-SADA reader) & 48.55 / 24.16 & 51.58 / 26.85 \\
Adapted & 40.38 / 21.31 & 42.51 / 23.34 \\
Nemotron 0.6B, fine-tuned on SADA & 46.50 / 21.21 & 48.22 / 22.62 \\
Cohere Transcribe Arabic 2B & 46.85 / 27.59 & 47.88 / 29.00 \\
\bottomrule\end{tabular}\end{table}

On SADA test the adapted system has the lowest WER of all systems: \R{ad.sada.clean16k.ctclm.wer}\% on clean audio, against \R{sada.cohere.clean16k.wer}\% for Cohere and \R{sada.nemo.clean16k.wer}\% for Nemotron. The adapted system also has the lowest WER on telephone-coded audio. This comparison favours the systems trained on SADA. Our readers and the fine-tuned Nemotron model saw the SADA training set, whereas we do not know whether Cohere did. We therefore treat SADA as an in-domain benchmark rather than a fair test of generalisation. Nemotron has a slightly lower CER than the adapted system on this set (\R{sada.nemo.clean16k.cer}\% against \R{ad.sada.clean16k.ctclm.cer}\% on clean audio).

\subsection{Further Human-Labelled Test Sets}
Table~\ref{tab:research} compares the telephone-SADA reader with Cohere Transcribe Arabic on the three further sets, and
Table~\ref{tab:boot} gives the paired differences.

\begin{table*}[t]\centering\small\caption{Additional Saudi and Gulf test sets: WER / CER (\%) of Cohere Transcribe Arabic and of the frozen-encoder system (telephone-SADA reader) with and without the language model.}\label{tab:research}
\resizebox{\textwidth}{!}{\begin{tabular}{lrccc}\toprule
Test set & $n$ & Cohere & Ours + LM & Ours greedy \\\midrule
Saudi customer-service calls (clean) & 48 & 19.27 / 10.05 & 29.62 / 11.91 & 34.50 / 11.07 \\
Saudi customer-service calls (telephone) & 48 & 20.53 / 10.92 & 30.63 / 12.72 & 37.04 / 12.13 \\
Casablanca, Emirati (clean) & 813 & 48.35 / 20.35 & 54.63 / 21.26 & 62.79 / 19.93 \\
Casablanca, Emirati (telephone) & 813 & 53.15 / 24.38 & 58.82 / 25.38 & 67.24 / 23.36 \\
SaudiTalk long-form (clean) & 133 & 56.75 / 29.51 & 65.29 / 29.60 & 70.17 / 28.61 \\
SaudiTalk long-form (telephone) & 133 & 55.80 / 28.40 & 66.56 / 31.36 & 72.46 / 30.47 \\
\bottomrule\end{tabular}}\end{table*}

\begin{table*}[t]\centering\small\caption{Paired difference, frozen-encoder system (+LM) minus Cohere, in percentage points, with 95\% bootstrap intervals over utterances (10{,}000 resamples); the differences are bootstrap means of per-utterance differences. Negative values favour the frozen-encoder system. The last column gives corpus-level WER (total errors over total reference words).}\label{tab:boot}
\resizebox{\textwidth}{!}{\begin{tabular}{lrccc}\toprule
Test set & $n$ & $\Delta$WER & $\Delta$CER & Corpus WER (Cohere / ours) \\\midrule
Saudi customer-service calls (clean) & 48 & +10.35 [+7.16, +13.50] & +1.85 [-0.44, +4.00] & 20.77 / 30.12 \\
Saudi customer-service calls (telephone) & 48 & +10.10 [+7.49, +12.79] & +1.81 [-0.06, +3.65] & 21.87 / 31.91 \\
Casablanca, Emirati (clean) & 813 & +6.28 [+4.52, +8.05] & +0.91 [-0.52, +2.30] & 47.52 / 53.53 \\
Casablanca, Emirati (telephone) & 813 & +5.67 [+3.74, +7.57] & +1.00 [-0.60, +2.59] & 51.75 / 57.32 \\
SaudiTalk long-form (clean) & 133 & +8.54 [+6.89, +10.20] & +0.09 [-1.58, +1.56] & 52.69 / 61.35 \\
SaudiTalk long-form (telephone) & 133 & +10.76 [+9.37, +12.14] & +2.96 [+1.96, +3.96] & 51.89 / 63.07 \\
\bottomrule\end{tabular}}\end{table*}

Cohere is better in WER on every set, and every WER interval excludes zero: for example by \R{boot.casablanca_uae.clean16k.wer} points on Emirati speech and by
\R{boot.nexdata_cs_sample.clean16k.wer} on the customer-service sample. Corpus-level WER gives the same ordering. At the character level the picture differs. On Casablanca the
CER gap is \R{boot.casablanca_uae.clean16k.cer} points (interval \R{boot.casablanca_uae.clean16k.cer.lo} to
\R{boot.casablanca_uae.clean16k.cer.hi}), and on SaudiTalk it is \R{boot.sauditalk_longform.clean16k.cer} (interval
\R{boot.sauditalk_longform.clean16k.cer.lo} to \R{boot.sauditalk_longform.clean16k.cer.hi}). Both intervals include zero, and so does the customer-service call's (\R{boot.nexdata_cs_sample.clean16k.cer} points, interval \R{boot.nexdata_cs_sample.clean16k.cer.lo} to \R{boot.nexdata_cs_sample.clean16k.cer.hi}): in CER the frozen system is not significantly different from Cohere on every condition in Table~\ref{tab:boot} except telephone-coded SaudiTalk, where it is \R{boot.sauditalk_longform.tel8k.cer} points worse. The frozen encoder therefore recovers most of the letters but fewer whole words. Two factors likely contribute: CER ignores word boundaries, which the 4-gram model only partly restores, and Cohere's decoder carries a much stronger implicit language model than our 4-gram.

On the three sets of Table~\ref{tab:research}, the language model lowers WER on every set but raises CER on every one of them. Without it, our reader's CER on Casablanca is already below Cohere's, at \R{res.v3_tel.casablanca_uae.clean16k.greedy.cer}\% against \R{res.cohere.casablanca_uae.clean16k.cer}\% on clean audio and \R{res.v3_tel.casablanca_uae.tel8k.greedy.cer}\% against \R{res.cohere.casablanca_uae.tel8k.cer}\% on telephone audio. On SaudiTalk it is \R{res.v3_tel.sauditalk_longform.clean16k.greedy.cer}\% against Cohere's \R{res.cohere.sauditalk_longform.clean16k.cer}\% on clean audio. On these sets, the 4-gram model thus repairs word boundaries but sometimes replaces a correctly spelled dialect word with a more frequent one. A user who needs letters rather than words is better served by greedy decoding.

The customer-service sample is the only public set in our domain, and it shows the largest WER gap. It contains a single speaker, however, so its intervals reflect only within-call variation. It should not be read as evidence about call centres in general.

\subsection{Adapting the Encoder}
\label{sec:res_adapted}
Table~\ref{tab:v5c} adds the adapted system. On the validation calls, whose labels are machine transcripts, the best greedy CTC WER rises from \R{ladder.chars_frozen}\% for the frozen character reader to \R{ladder.subword_lora}\% for the sub-word reader with attention adapters after two epochs. It then falls to \R{ladder.hybrid}\% for the hybrid reader, and to \R{ladder.hybrid_selfcond}\% after the joint stage. Adding the feed-forward and convolution adapters brings it to \R{ladder.ffn_lora}\%. These stages differ in training length and data, so they form a development history rather than a controlled ablation.

\begin{table*}[t]\centering\small\caption{Human-labelled test sets, WER / CER (\%): Cohere Transcribe Arabic, the frozen system (telephone-SADA reader, word 4-gram) and the adapted system with CTC beam search and a word 4-gram, or with joint CTC/attention search (beam 16) and a sub-word 6-gram; Nemotron is the 0.6B model fine-tuned on SADA. A dash marks a run not made.}\label{tab:v5c}
\resizebox{\textwidth}{!}{\begin{tabular}{lccccc}\toprule
& & & Frozen & \multicolumn{2}{c}{Adapted}\\
Test set & Cohere & Nemotron & CTC + LM & CTC + LM & Joint + LM \\\midrule
Saudi customer-service calls (clean) & 19.27 / 10.05 & 24.12 / 8.58 & 29.62 / 11.91 & 20.86 / 10.02 & 19.20 / 7.77 \\
Saudi customer-service calls (telephone) & 20.53 / 10.92 & 25.41 / 9.85 & 30.63 / 12.72 & 22.61 / 10.58 & 21.02 / 8.39 \\
Casablanca, Emirati (clean) & 48.35 / 20.35 & 53.27 / 17.09 & 54.63 / 21.26 & 50.66 / 20.77 & 46.98 / 14.89 \\
Casablanca, Emirati (telephone) & 53.15 / 24.38 & 56.56 / 19.81 & 58.82 / 25.38 & 53.50 / 23.53 & 50.01 / 17.40 \\
SaudiTalk long-form (clean) & 56.75 / 29.51 & 61.25 / 25.76 & 65.29 / 29.60 & 60.29 / 29.78 & 58.79 / 25.58 \\
SaudiTalk long-form (telephone) & 55.80 / 28.40 & 62.63 / 27.22 & 66.56 / 31.36 & -- & 60.45 / 26.65 \\
\bottomrule\end{tabular}}\end{table*}

\begin{figure*}[t]\centering\includegraphics[width=\textwidth]{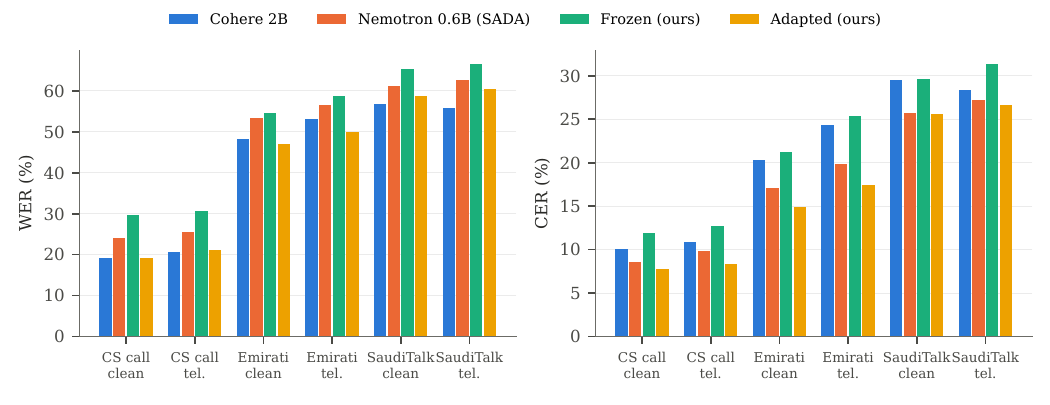}
\caption{WER and CER on the human-labelled test sets for Cohere, the SADA-fine-tuned Nemotron model and our two systems (adapted: joint search; values in Table~\ref{tab:v5c}).}\label{fig:results}\end{figure*}

Figure~\ref{fig:results} summarises these results.

On the human-labelled sets, the adapted system lowers WER relative to the frozen system on every set, and joint decoding lowers it further on every set where both searches were run. Table~\ref{tab:boot5} compares it with Cohere. On the customer-service call, joint decoding gives \R{ad.nexdata_cs_sample.clean16k.jointlm.wer}\% WER against Cohere's \R{res.cohere.nexdata_cs_sample.clean16k.wer}\%, a difference of \R{boot5.nexdata_cs_sample.clean16k.wer} points (interval \R{boot5.nexdata_cs_sample.clean16k.wer.lo} to \R{boot5.nexdata_cs_sample.clean16k.wer.hi}), and \R{boot5.nexdata_cs_sample.tel8k.wer} points on its telephone version. The two systems are not significantly different in WER there. Its CER is significantly lower: \R{ad.nexdata_cs_sample.clean16k.jointlm.cer}\% against \R{res.cohere.nexdata_cs_sample.clean16k.cer}\% (difference \R{boot5.nexdata_cs_sample.clean16k.cer}, interval \R{boot5.nexdata_cs_sample.clean16k.cer.lo} to \R{boot5.nexdata_cs_sample.clean16k.cer.hi}), and \R{ad.nexdata_cs_sample.tel8k.jointlm.cer}\% against \R{res.cohere.nexdata_cs_sample.tel8k.cer}\% on telephone audio. On Emirati speech, the WER difference is \R{boot5.casablanca_uae.clean16k.wer} points on clean audio (interval \R{boot5.casablanca_uae.clean16k.wer.lo} to \R{boot5.casablanca_uae.clean16k.wer.hi}, not significant) and \R{boot5.casablanca_uae.tel8k.wer} on telephone-coded audio (interval \R{boot5.casablanca_uae.tel8k.wer.lo} to \R{boot5.casablanca_uae.tel8k.wer.hi}), where the adapted system is significantly better. Its CER differs by \R{boot5.casablanca_uae.clean16k.cer} and \R{boot5.casablanca_uae.tel8k.cer} points (negative favours the adapted system). On SaudiTalk, Cohere remains ahead in WER by \R{boot5.sauditalk_longform.clean16k.wer} points on clean and \R{boot5.sauditalk_longform.tel8k.wer} points on telephone-coded audio, both significant, while the adapted system's CER is significantly lower (differences \R{boot5.sauditalk_longform.clean16k.cer} and \R{boot5.sauditalk_longform.tel8k.cer} points). Joint decoding contributes much of the CER gain. On Emirati speech it lowers CER from \R{ad.casablanca_uae.clean16k.ctclm.cer}\% with CTC search to \R{ad.casablanca_uae.clean16k.jointlm.cer}\%.

\begin{table*}[t]\centering\small\caption{Paired difference, adapted system minus Cohere, in percentage points, with 95\% bootstrap intervals over utterances (10{,}000 resamples). Negative values favour the adapted system. The last column gives corpus-level WER (Cohere / adapted).}\label{tab:boot5}
\resizebox{\textwidth}{!}{\begin{tabular}{llccc}\toprule
Test set & Search & $\Delta$WER & $\Delta$CER & Corpus WER \\\midrule
Saudi customer-service calls (clean) & Joint + LM & -0.07 [-3.90, +3.48] & -2.28 [-4.45, -0.21] & 20.77 / 19.81 \\
Saudi customer-service calls (telephone) & Joint + LM & +0.49 [-2.51, +3.46] & -2.53 [-4.23, -0.84] & 21.87 / 21.60 \\
Casablanca, Emirati (clean) & Joint + LM & -1.37 [-3.11, +0.43] & -5.47 [-6.88, -4.10] & 47.52 / 47.38 \\
Casablanca, Emirati (telephone) & Joint + LM & -3.14 [-5.01, -1.32] & -6.98 [-8.63, -5.40] & 51.75 / 50.06 \\
SaudiTalk long-form (clean) & Joint + LM & +2.04 [+0.64, +3.41] & -3.93 [-5.59, -2.49] & 52.69 / 54.34 \\
SaudiTalk long-form (telephone) & Joint + LM & +4.64 [+3.41, +5.86] & -1.76 [-2.85, -0.73] & 51.89 / 56.20 \\
\bottomrule\end{tabular}}\end{table*}

Table~\ref{tab:bootn} compares the adapted system with the Nemotron model fine-tuned on SADA. Its telephone fine-tuning used GSM coding, whereas our telephone-coded test versions use G.711 $\mu$-law, a mismatch that may slightly disadvantage it on the telephone conditions. The adapted system has a significantly lower WER on the customer-service call, by \R{bootn.nexdata_cs_sample.clean16k.wer} points (interval \R{bootn.nexdata_cs_sample.clean16k.wer.lo} to \R{bootn.nexdata_cs_sample.clean16k.wer.hi}). It is also significantly lower on Emirati speech, by \R{bootn.casablanca_uae.clean16k.wer} and \R{bootn.casablanca_uae.tel8k.wer} points, and its CER on Emirati speech is lower too. On SaudiTalk its WER is significantly lower, by \R{bootn.sauditalk_longform.clean16k.wer} and \R{bootn.sauditalk_longform.tel8k.wer} points. The CERs are close, differing by \R{bootn.sauditalk_longform.clean16k.cer} points (interval \R{bootn.sauditalk_longform.clean16k.cer.lo} to \R{bootn.sauditalk_longform.clean16k.cer.hi}) on clean audio.

\begin{table*}[t]\centering\small\caption{Paired difference, adapted system minus Nemotron 0.6B (fine-tuned on SADA), in percentage points, with 95\% bootstrap intervals over utterances. Negative values favour the adapted system.}\label{tab:bootn}
\resizebox{\textwidth}{!}{\begin{tabular}{llcc}\toprule
Test set & Search & $\Delta$WER & $\Delta$CER \\\midrule
Saudi customer-service calls (clean) & Joint + LM & -4.92 [-8.25, -1.68] & -0.81 [-2.05, +0.38] \\
Saudi customer-service calls (telephone) & Joint + LM & -4.39 [-7.81, -1.07] & -1.46 [-2.88, -0.12] \\
Casablanca, Emirati (clean) & Joint + LM & -6.28 [-7.70, -4.88] & -2.20 [-2.92, -1.50] \\
Casablanca, Emirati (telephone) & Joint + LM & -6.55 [-8.00, -5.13] & -2.41 [-3.17, -1.63] \\
SaudiTalk long-form (clean) & Joint + LM & -2.46 [-3.35, -1.59] & -0.19 [-0.59, +0.23] \\
SaudiTalk long-form (telephone) & Joint + LM & -2.18 [-3.02, -1.33] & -0.58 [-1.00, -0.15] \\
\bottomrule\end{tabular}}\end{table*}

The 16-best lists from the joint search show how much room remains. On the customer-service call, the best of the 16 hypotheses scores \R{ad.nb16.nexdata_cs_sample.clean16k.oracle}\% WER, or \R{ad.nb16.nexdata_cs_sample.tel8k.oracle}\% on telephone audio. This is well below both the top hypothesis and Cohere. On Emirati speech it scores \R{ad.nb16.casablanca_uae.clean16k.oracle}\% and \R{ad.nb16.casablanca_uae.tel8k.oracle}\%. The adapted system already produces many transcripts better than the 2B recogniser's, unlike the frozen system (Section~\ref{sec:search}), so better ranking could still lower its WER. These are oracle bounds, not achieved results.

\subsection{Blind Call-Centre Sample}
Table~\ref{tab:dataocean} adds a third drafting system, the commercial ElevenLabs Scribe v2 \cite{scribe_v2}, on the blind call-centre sample. The adapted system has the lowest WER and CER on clean audio and the lowest CER on telephone-coded audio, where Cohere's WER is slightly lower. The sample is small, however: in paired comparisons on clips drafted by neither system, the WER difference (adapted minus other; negative favours the adapted system) is \R{api.cohere.d} points against Cohere (interval \R{api.cohere.d.lo} to \R{api.cohere.d.hi}, \R{api.cohere.d.n} clips) and \R{api.scribev2.d} points against Scribe (interval \R{api.scribev2.d.lo} to \R{api.scribev2.d.hi}), and among all these comparisons only the CER difference against Scribe on clean audio (\R{dob.scribe_v2.clean16k.cer} points, interval \R{dob.scribe_v2.clean16k.cer.lo} to \R{dob.scribe_v2.clean16k.cer.hi}) is statistically significant. The direction agrees with the other call-centre results, but the sample is too small to establish a difference on its own. The annotator checked every clip against the audio and, after normalisation, accepted unchanged \R{do.unchanged.scribe} of \R{do.drafted.scribe} Scribe drafts, \R{do.unchanged.ours} of \R{do.drafted.ours} drafts from our system and \R{do.unchanged.cohere} of \R{do.drafted.cohere} Cohere drafts. Scribe's median clip WER (\R{do.median.scribe_v2}\%) is no higher than the others' (\R{do.median.v5d_b16}\% for ours, \R{do.median.cohere}\% for Cohere), but it fails completely, with a WER of 100\% or more, on \R{do.fail100.scribe_v2} clips against \R{do.fail100.v5d_b16} and \R{do.fail100.cohere}, which raises its mean. Excluding each system's own drafts removes the direct advantage of drafting; a second, independent annotator would be needed to rule out any remaining effect of the drafts on the references.

\begin{table*}[t]\centering\small\caption{Public Saudi call-centre dialogue sample with human-corrected references and a balanced blind prefill: each clip's draft came from one of the three systems, hidden from the annotator, and each system is scored only on clips it did not draft (number of clips in brackets). WER / CER (\%).}\label{tab:dataocean}
\begin{tabular}{lcc}\toprule
System & Clean & Telephone \\\midrule
Adapted (ours) & 29.89 / 11.30 (61) & 31.51 / 11.92 (61) \\
Cohere Transcribe Arabic 2B & 34.17 / 14.59 (62) & 30.87 / 14.10 (62) \\
ElevenLabs Scribe v2 & 35.32 / 17.16 (61) & 37.00 / 19.30 (61) \\
\bottomrule\end{tabular}\end{table*}

Because all \R{api.n_systems} systems can be scored on the clips not drafted by our system, the blind sample also allows a wider comparison with commercial speech-recognition services, accessed through OpenRouter \cite{openrouter} with the language set to Arabic (Table~\ref{tab:apis}). Three commercial APIs score better than the adapted system, significantly so only GPT-Transcribe: GPT-Transcribe \cite{api_gpt_transcribe} is clearly better (\R{api.gpttranscribe.d} WER points in our disfavour, interval \R{api.gpttranscribe.d.lo} to \R{api.gpttranscribe.d.hi}), and Chirp~3 \cite{api_chirp3} and MAI-Transcribe-2 \cite{api_mai_transcribe2} are ahead by smaller, non-significant margins. The adapted system ranks \R{api.rank_ours} of \R{api.n_systems} and has the lowest WER of the self-hostable systems tested by point estimate (its gap to Cohere is not significant); it is significantly better than Whisper large-v3, both Voxtral models, Deepgram Nova-3 \cite{api_deepgram}, GPT-4o Transcribe \cite{api_gpt4o_transcribe} and Fish Audio \cite{api_fish}, although the intervals for Deepgram Nova-3 (upper bound \R{api.nova3.d.hi}) and Fish Audio (\R{api.transcribe1pro.d.hi}) only just exclude zero, and not significantly different from Gemini 3.5 Transcribe \cite{api_gemini_transcribe}, Grok STT \cite{api_grok_stt}, AssemblyAI \cite{api_assemblyai}, Cohere and Scribe. These comparisons rest on \R{api.gpttranscribe.d.n} clips and one annotator, so they indicate rankings rather than precise gaps. Deepgram, the vendor of Nova-3, also supplied the cloud recogniser that produced half of our call-centre training labels (Section~\ref{sec:setup}), so the comparison with Nova-3 is not independent of our training data.

\begin{table*}[t]\centering\small\caption{Blind call-centre sample: 15 recognisers sorted by clean WER. Each system is scored on the clips that neither it nor our system drafted ($n$); commercial APIs drafted none. The last column is the paired WER difference, ours minus that system, on clean audio over clips drafted by neither, with a 95\% bootstrap interval; $^{*}$ marks intervals that exclude zero. WER / CER (\%).}\label{tab:apis}
\resizebox{\textwidth}{!}{\begin{tabular}{lcrccc}\toprule
System & Self-hostable & $n$ & Clean & Telephone & $\Delta$WER vs ours \\\midrule
GPT-Transcribe \cite{api_gpt_transcribe} & no & 61 & 18.84 / 8.14 & 19.21 / 9.13 & +11.05 [+6.14, +16.17]$^{*}$ \\
Chirp 3 \cite{api_chirp3} & no & 61 & 26.50 / 9.78 & 25.70 / 10.12 & +3.39 [-1.48, +8.17] \\
MAI-Transcribe-2 \cite{api_mai_transcribe2} & no & 61 & 27.28 / 9.12 & 24.32 / 9.11 & +2.61 [-1.88, +6.63] \\
Adapted (ours) & yes & 61 & 29.89 / 11.30 & 31.51 / 11.92 & -- \\
Grok STT 1.0 \cite{api_grok_stt} & no & 61 & 31.23 / 17.12 & 31.94 / 14.25 & -1.34 [-8.10, +4.94] \\
Gemini 3.5 Transcribe \cite{api_gemini_transcribe} & no & 61 & 31.33 / 13.05 & 29.73 / 11.62 & -1.44 [-8.09, +4.66] \\
AssemblyAI Universal-3.5 Pro \cite{api_assemblyai} & no & 61 & 32.15 / 12.83 & 32.80 / 13.89 & -2.26 [-9.31, +4.17] \\
Cohere Transcribe Arabic 2B \cite{cohere_transcribe} & yes & 31 & 35.06 / 12.11 & 33.07 / 14.63 & -5.47 [-12.45, +1.69] \\
Deepgram Nova-3 \cite{api_deepgram} & no & 61 & 36.03 / 13.45 & 36.97 / 16.22 & -6.14 [-12.61, -0.14]$^{*}$ \\
Voxtral Mini Transcribe \cite{api_voxtral} & no & 61 & 38.29 / 16.64 & 39.16 / 17.91 & -8.40 [-15.54, -2.17]$^{*}$ \\
GPT-4o Transcribe \cite{api_gpt4o_transcribe} & no & 61 & 38.65 / 27.63 & 44.83 / 30.54 & -8.76 [-15.98, -1.83]$^{*}$ \\
Fish Audio Transcribe 1 Pro \cite{api_fish} & no & 61 & 39.50 / 21.11 & 38.61 / 21.09 & -9.61 [-19.27, -0.91]$^{*}$ \\
ElevenLabs Scribe v2 \cite{scribe_v2} & no & 30 & 39.71 / 15.02 & 39.38 / 17.39 & -9.50 [-20.10, +0.25] \\
Voxtral Small 24B \cite{api_voxtral} & yes & 61 & 39.74 / 16.71 & 44.86 / 20.53 & -9.85 [-18.27, -2.23]$^{*}$ \\
Whisper large-v3 \cite{whisper} & yes & 61 & 44.35 / 22.84 & 46.74 / 22.57 & -14.46 [-22.21, -7.08]$^{*}$ \\
\bottomrule\end{tabular}}\end{table*}

\begin{figure}[t]\centering\includegraphics[width=\columnwidth]{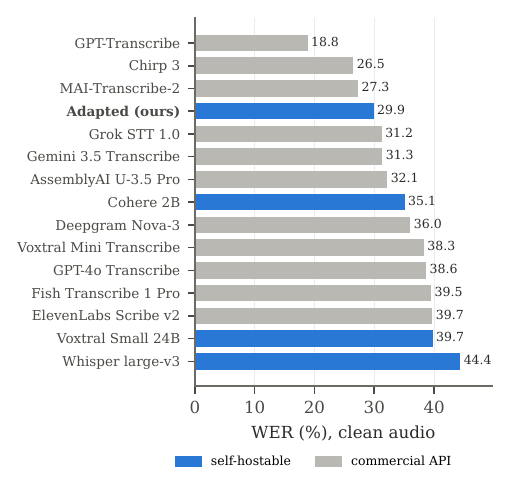}
\caption{Clean-audio WER of the fifteen systems on the blind call-centre sample, sorted; self-hostable systems in blue. Values and intervals in Table~\ref{tab:apis}.}\label{fig:blind}\end{figure}

Figure~\ref{fig:blind} shows the ranking.

\begin{figure}[t]\centering\includegraphics[width=\columnwidth]{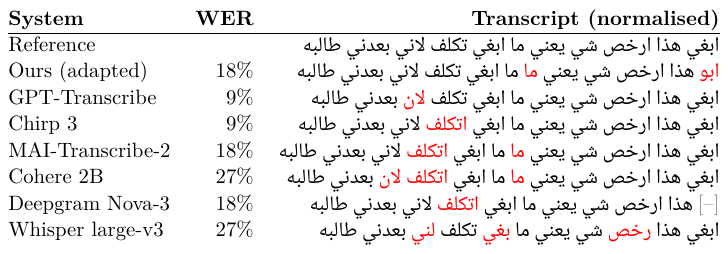}
\caption{Outputs of seven systems for one clip of the blind call-centre sample, after the scoring normalisation. Red words differ from the reference; a grey placeholder marks a missing reference word.}\label{fig:example}\end{figure}

Figure~\ref{fig:example} shows the outputs of several systems for one clip. The clip was chosen by a fixed rule rather than by hand: among clips drafted by Scribe (which is therefore not shown) with 7 to 14 reference words, it is the one whose mean WER across the shown systems is closest to the median. Most systems differ only in one or two words, typically a dialect form or a spelling of the same word.

Table~\ref{tab:final} brings accuracy, deployment and cost together. The commercial systems that are more accurate cost between \$\R{price.better.min} and \$\R{price.better.max} per audio hour, and they send audio to a third party. The adapted system runs on one GPU at about \R{fast.p4.rtfx} times real time. On the hardware we used, thousand hours of audio take about \R{fast.gpuh_per_1000h} GPU-hours to process.

\begin{table*}[t]\centering\small\caption{Final comparison on the blind call-centre sample (clean audio, clips drafted by neither the system nor ours). Speed is end-to-end throughput measured on one Quadro RTX 6000 (Table~\ref{tab:speed2}; different audio and optimisation level; see Section~\ref{sec:fast}); cloud APIs were not timed comparably. Price is the list price for batch transcription per audio hour, fetched on 2026-10-16 from each vendor's pricing page (MAI-Transcribe-2 via OpenRouter, for which no vendor price is published); for open-weight models we accessed through an API, the API price is shown. Self-hosted systems pay only for their own hardware.}\label{tab:final}
\resizebox{\textwidth}{!}{\begin{tabular}{lcccc}\toprule
System & Self-hostable & WER / CER (\%) & Speed & Price per hour \\\midrule
GPT-Transcribe & no & 18.84 / 8.14 & cloud API & \$0.27 \\
Chirp 3 & no & 26.50 / 9.78 & cloud API & \$0.96 \\
MAI-Transcribe-2 & no & 27.28 / 9.12 & cloud API & \$0.10 \\
Adapted (ours) & yes & 29.89 / 11.30 & 139$\times$ & self-hosted \\
Grok STT 1.0 & no & 31.23 / 17.12 & cloud API & \$0.10 \\
Gemini 3.5 Transcribe & no & 31.33 / 13.05 & cloud API & \$0.30 \\
AssemblyAI Universal-3.5 Pro & no & 32.15 / 12.83 & cloud API & \$0.21 \\
Cohere Transcribe Arabic 2B & yes & 35.06 / 12.11 & 80$\times$ & self-hosted \\
Deepgram Nova-3 & no & 36.03 / 13.45 & cloud API & \$0.26 \\
Voxtral Mini Transcribe & no & 38.29 / 16.64 & cloud API & \$0.18 \\
GPT-4o Transcribe & no & 38.65 / 27.63 & cloud API & \$0.36 \\
Fish Audio Transcribe 1 Pro & no & 39.50 / 21.11 & cloud API & \$0.36 \\
ElevenLabs Scribe v2 & no & 39.71 / 15.02 & cloud API & \$0.22 \\
Voxtral Small 24B & yes & 39.74 / 16.71 & not measured & \$0.24 \\
Whisper large-v3 & yes & 44.35 / 22.84 & not measured & \$0.03 \\
\bottomrule\end{tabular}}\end{table*}

\subsection{Call-Centre Recordings}
Table~\ref{tab:gold} gives results on the in-house calls. With the language model, the telephone-SADA reader reaches \R{gold.v3_tel.lm.wer}\% WER and \R{gold.v3_tel.lm.cer}\% CER. The fine-tuned Nemotron model reaches \R{gold.nemo_ft.wer}\% WER and a clearly better \R{gold.nemo_ft.cer}\% CER. The WER difference, \R{gold.diffnemo} points, is not significant (window bootstrap \R{gold.diffnemo.lo} to \R{gold.diffnemo.hi}; call bootstrap \R{gold.diffnemo.clo} to \R{gold.diffnemo.chi}), but a non-significant difference on \R{gold.n} windows does not establish equivalence, and the two systems were trained on different data: our reader saw in-domain calls, while Nemotron saw only SADA. Cohere scores \R{gold.cohere.wer}\%, a difference of \R{gold.diff} points (window bootstrap \R{gold.diff.lo} to \R{gold.diff.hi}; call bootstrap \R{gold.diff.clo} to \R{gold.diff.chi}). Corpus-level WER is \R{gold.v3_tel.lm.corpus_wer}\% for our reader, \R{gold.nemo_ft.corpus_wer}\% for Nemotron and \R{gold.cohere.corpus_wer}\% for Cohere. The zero-shot Nemotron model scores \R{gold.nemo_base.wer}\%. The adapted system reaches \R{ad.gold.ctclm.wer}\% WER with CTC search and \R{ad.gold.jointlm.wer}\% WER (\R{ad.gold.jointlm.cer}\% CER) with joint search.

\begin{table}[t]\centering\small\caption{In-house call-centre recordings (72 windows, 8~kHz; machine-drafted references reviewed by one annotator, see Section~\ref{sec:setup}): mean per-window WER and CER (\%) with greedy CTC decoding and with the 4-gram language model. The reference systems use their own decoders, so their scores appear in the right-hand columns.}\label{tab:gold}
\begin{tabular}{lrrrr}\toprule
& \multicolumn{2}{c}{Greedy} & \multicolumn{2}{c}{+ 4-gram LM}\\
System & WER & CER & WER & CER \\\midrule
Calls only & 60.43 & 27.10 & 49.00 & 25.99 \\
+ SADA, telephone-coded & 58.69 & 25.61 & 47.20 & 25.32 \\
+ SADA, clean & 58.15 & 25.65 & 47.26 & 25.47 \\
\midrule
Nemotron 3.5 ASR 0.6B, zero-shot (beam 8) & \multicolumn{2}{c}{--} & 51.33 & 25.83 \\
Nemotron 3.5 ASR 0.6B, fine-tuned on SADA (beam 8) & \multicolumn{2}{c}{--} & 46.71 & 21.47 \\
Cohere Transcribe Arabic 2B (greedy) & \multicolumn{2}{c}{--} & 37.62 & 19.67 \\
\bottomrule\end{tabular}\end{table}

\begin{table}[t]\centering\small\caption{Call-centre WER (\%) split by where the reference came from: Cohere draft accepted unchanged (30 windows), commercial cloud recogniser's draft accepted unchanged (33), and edited by the annotator (9).}\label{tab:provenance}
\begin{tabular}{lrrr}\toprule
System & Cohere draft & Cloud ASR draft & Edited \\\midrule
Calls only & 60.11 & 37.87 & 52.75 \\
+ SADA, telephone-coded & 58.38 & 36.28 & 50.01 \\
+ SADA, clean & 57.54 & 37.14 & 50.11 \\
Nemotron 0.6B, fine-tuned & 56.16 & 35.94 & 54.72 \\
Cohere 2B & 43.60 & 30.88 & 42.37 \\
\bottomrule\end{tabular}\end{table}

Table~\ref{tab:provenance} shows why the Cohere comparison on this set must be read with care. On the \R{gold.src.n.coh} windows whose reference is an unchanged Cohere draft, Cohere leads the frozen-encoder system by \R{gold.prov.coh.gap} points. On the \R{gold.src.n.dg} windows whose reference is an unchanged draft from the commercial cloud recogniser, the lead shrinks to \R{gold.prov.dg.gap} points (\R{gold.cohere.src.dg.wer}\% against \R{gold.v3_tel.lm.src.dg.wer}\%). Part of Cohere's advantage on the full set is therefore an artefact of how the references were made. The subset drafted by the commercial cloud recogniser is not neutral either, since that system also labelled our training data.

Table~\ref{tab:category} separates conversational windows from windows dominated by spoken numbers. For every system, WER on the two kinds differs by at most about two points. Spoken digit strings are therefore not a particular weakness of the frozen encoder relative to the other systems.

\begin{table}[t]\centering\small\caption{Call-centre WER (\%) by window type: conversational speech (43 windows) and windows dominated by spoken numbers such as identity, phone and account numbers (29).}\label{tab:category}
\begin{tabular}{lrr}\toprule
System & Speech & Numbers \\\midrule
Calls only & 49.25 & 48.63 \\
+ SADA, telephone-coded & 47.70 & 46.47 \\
+ SADA, clean & 47.16 & 47.41 \\
Nemotron 0.6B, fine-tuned & 47.41 & 45.67 \\
Cohere 2B & 37.36 & 38.00 \\
\bottomrule\end{tabular}\end{table}

\subsection{Which Layers Does the Reader Use?}
\begin{figure}[t]\centering\includegraphics[width=0.7\columnwidth]{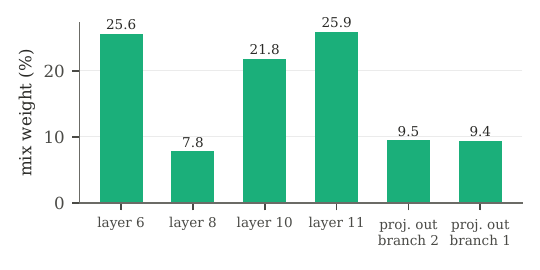}
\caption{Learned layer-mix weights of the frozen telephone-SADA reader. The last two bars are two branches that both carry the projected encoder output (Section~\ref{sec:frozenenc}).}\label{fig:layers}\end{figure}

Figure~\ref{fig:layers} plots the weights. The learned weights of Eq.~\eqref{eq:mix} for the telephone-SADA reader are \R{mix.v3_tel.6}\% for layer 6, \R{mix.v3_tel.8}\% for layer 8, \R{mix.v3_tel.10}\% for layer 10, \R{mix.v3_tel.11}\% for layer 11, and \R{mix.v3_tel.12}\% and \R{mix.v3_tel.out}\% for the two branches that both carry the projected output (Section~\ref{sec:frozenenc}), a combined share of 18.9\% for that representation. Layers 6, 10 and 11 together receive more than two thirds of the weight. These weights should not be read as layer importance. Each branch has its own normalisation and projection, and these can absorb a scale factor. Only these five representations were offered to the mix, and the raw output of layer 12 was not among them, so a single-layer probe of all twelve layers is needed before any conclusion is drawn about where the encoder holds transcription-relevant detail.

\subsection{Search or Model?}
\label{sec:search}
Could better re-ranking close the call-centre gap? We computed the oracle WER of a wider beam (128 hypotheses, same language model), that is, the lowest WER among the top $N$ hypotheses of each window. The top hypothesis of this wider search scores \R{oracle.top1.beam128}\%, slightly different from the \R{gold.v3_tel.lm.wer}\% of the beam-64 system. The oracle falls to \R{oracle.lm.top8}\% for the top 8 and \R{oracle.lm.top128}\% for the top 128, and pooling the beams with and without the language model gives \R{oracle.union}\%. Even a perfect re-ranker would stay above Cohere's \R{gold.cohere.wer}\%. The same holds on the \R{gold.src.n.dg} windows with cloud-recogniser references, which Cohere did not draft: there the top-128 oracle is \R{oracle.top128.src.dg}\%, against \R{gold.cohere.src.dg.wer}\% for Cohere. The gap therefore lies in the acoustic model and its candidates, not in search.

\subsection{Speed and Compute Cost}
\label{sec:speed}
Table~\ref{tab:speed} reports throughput measured on identical audio, one system at a time on an otherwise idle GPU, with batches
sorted by length so that no system is penalised by padding; on the GPU both systems use batches of 16, and Cohere is also shown with 32. All systems run in PyTorch without further inference optimisation. The encoder on its own processes audio at
\R{speed.eg2.enc_rtfx} times real time on the GPU; the complete pipeline with greedy decoding, including feature extraction and the reader, runs at \R{speed.eg2.gpu_rtfx} times real time and needs \R{speed.eg2.gpu_mem}~GiB of GPU memory. At the same batch size,
Cohere Transcribe Arabic runs at \R{speed.cohere.gpu_rtfx16} times real time with \R{speed.cohere.gpu_mem}~GiB, so the frozen-encoder
system is \R{speed.ratio.gpu16} times faster and uses \R{speed.ratio.mem} times less memory. The speed advantage on the GPU is
modest, however: with a larger batch Cohere reaches \R{speed.cohere.gpu_rtfx32} times real time, because its autoregressive decoder
produces few tokens for short utterances and parallelises well.

On CPU, often the deployment target in call centres, the difference is larger. With 16 threads, the frozen-encoder system runs at \R{speed.eg2.cpu_rtfx} times real time with greedy decoding, which is \R{speed.ratio.cpu} times faster than Cohere (\R{speed.cohere.cpu_rtfx}) and \R{speed.ratio.cpu_nemo} times faster than the 0.6B Nemotron model with greedy decoding (\R{speed.nemo.cpu_rtfx}). The accuracy figures above rely on 4-gram beam search, which adds work on the CPU: one process decodes at \R{speed.lm.cpu1_rtfx} times real time, or about \R{speed.lm.corehours} core-hours per 1,000 hours of audio, though utterances can be decoded in parallel. Nemotron's batched GPU inference failed with a memory-access error on the audio of Table~\ref{tab:speed}, so only its CPU figures appear there; it was timed on the GPU on different audio in Section~\ref{sec:fast}. These measurements cover 13 minutes of short utterances on CPU and one GPU type. They describe the frozen system with greedy decoding only. The adapted system is timed in Section~\ref{sec:fast}.

\begin{table*}[t]\centering\small\caption{Speed and compute cost on identical audio with length-sorted batches. RTFx is seconds of audio processed per second of wall time, including feature extraction and greedy decoding. GPU: one NVIDIA Quadro RTX 6000 (24~GB), 56 minutes of audio (Casablanca Emirati test). CPU: 16 threads of an Intel Xeon Gold 6230, 13 minutes of audio, batch 8, 32-bit floats. Cost is given as GPU-hours or CPU core-hours per 1{,}000 hours of audio. The 4-gram beam search adds 12 CPU core-hours per 1{,}000 hours (one process, RTFx 84).}\label{tab:speed}
\resizebox{\textwidth}{!}{\begin{tabular}{lrrrrrr}\toprule
& & \multicolumn{3}{c}{GPU} & \multicolumn{2}{c}{CPU (16 threads)} \\
System & Params (M) & RTFx & Mem. (GiB) & GPU-h & RTFx & core-h \\\midrule
Frozen encoder + reader (greedy) & 328 & 257 & 1.5 & 3.9 & 31.1 & 514 \\
\quad encoder only & 305 & 348 & -- & 2.9 & 34.4 & -- \\
Cohere Transcribe Arabic (greedy), batch 16 & 2,066 & 157 & 4.8 & 6.4 & 9.2 & 1,739 \\
Cohere Transcribe Arabic (greedy), batch 32 & 2,066 & 222 & 5.2 & 4.5 & -- & -- \\
Nemotron 0.6B fine-tuned (greedy) & 638 & -- & -- & -- & 22.6 & 708 \\
Nemotron 0.6B fine-tuned (beam 8) & 638 & -- & -- & -- & 19.4 & 825 \\
\bottomrule\end{tabular}}\end{table*}

\subsection{Fast Joint Decoding}
\label{sec:fast}
Joint CTC/attention search is what gives the adapted system its accuracy, but our first implementation ran at only \R{ref.joint_orig} times real time, because every search step issued many small GPU operations and scored CTC prefixes frame by frame in Python. We rewrote it in six steps, checking after each step how many transcripts changed: CTC prefix scores are computed by a compiled routine; the language model is queried only for the candidates that survive the attention pre-beam; the attention decoder keeps a key/value cache and projects the encoder memory once per utterance; one decoder call serves all utterances of a batch at every step; the decoder runs in 16-bit floating point, which the GPU executes natively, instead of bfloat16, which it emulates; and the encoder processes length-sorted batches rather than one segment at a time. Batching the search across utterances left all 60 checked transcripts unchanged, and half precision changed two of 198.

\begin{table}[t]\centering\small\caption{Throughput of the adapted system and the reference recognisers on one idle Quadro RTX 6000 (RTFx = seconds of audio per second of wall time, end to end). Upper part: the optimised joint CTC/attention pipeline with the sub-word LM (827 s of audio, batches of 16 utterances) and its WER/CER on the telephone customer-service call. Lower part: the original joint search and the other systems on the 1673 s of the in-house call windows.}\label{tab:speed2}
\begin{tabular}{lrc}\toprule
System & RTFx & CS-call tel. WER / CER \\\midrule
Optimised joint search, 1 process & 68 & 20.32 / 8.34 \\
Optimised joint search, 2 processes & 106 & 20.48 / 8.38 \\
Optimised joint search, 3 processes & 133 & 20.48 / 8.38 \\
Optimised joint search, 4 processes & 139 & 20.48 / 8.38 \\
\midrule
Original joint search (beam 10) & 1.7 & \\
Adapted, CTC greedy (batch 1 / 16) & 174 / 141 & \\
Adapted, CTC + word LM (12 CPU processes) & 136 & \\
Cohere Transcribe Arabic 2B (batch 1 / 16) & 15 / 80 & \\
Nemotron 0.6B (batch 1 / 16) & 131 / 183 & \\
\bottomrule\end{tabular}\end{table}

Table~\ref{tab:speed2} gives end-to-end throughput, from audio to text, on one idle GPU. With one process the optimised pipeline runs at \R{fast.p1.rtfx} times real time, and with four processes sharing the GPU at \R{fast.p4.rtfx} times, against \R{ref.joint_orig} times for the original search on different audio; WER on the telephone customer-service call moves from \R{fast.p1.cswer}\% to \R{fast.p4.cswer}\%. For comparison, CTC search with the word language model runs at \R{ref.ctclm} times real time, Cohere at \R{ref.cohere_b16} times with batches of 16 (higher in Table~\ref{tab:speed}, which used shorter audio), and Nemotron at \R{ref.nemo_b16} times. The joint search is therefore no longer the slow path. The two halves of the table were measured on different audio, so they are not directly comparable; on that basis the adapted system's 139-fold throughput is likely, but not shown, to exceed Cohere's on this GPU. The optimised pipeline differs slightly from the search used for the accuracy tables, which ran one utterance at a time in bfloat16 with a beam of 16; batching and half precision change a few transcripts, which is why WER differs slightly between Table~\ref{tab:v5c} and Table~\ref{tab:speed2}. Merging the adapters into the encoder weights changed a few transcripts for little speed gain, so we kept them separate.

\section{Discussion}
\label{sec:discussion}

\paragraph{What the frozen encoder can do} The speech encoder that Gemma 4 and EmbeddingGemma 2 share stays frozen and is never adapted to Arabic transcription. A small reader on top of it reaches a CER not significantly different from that of a 2B dedicated recogniser on five of the six human-labelled conditions we tested (Table~\ref{tab:boot}). On real call-centre audio, its WER is not significantly different from that of a 0.6B recogniser fine-tuned on SADA, although its CER is clearly worse. Only 22.6M parameters are trained, so adapting to a new domain is cheap. A deployment that already runs Gemma 4 or EmbeddingGemma 2 can add transcription without loading a second large encoder. The system is also cheaper to run. It is about \R{speed.ratio.cpu} times faster than the 2B recogniser on CPU, with \R{speed.ratio.mem} times less GPU memory.

\paragraph{What adaptation adds} With 3.1M added adapter parameters, sub-word units, an attention decoder and joint decoding, the system is not significantly different from the 2B recogniser in WER on the customer-service call and on clean Emirati speech. On telephone-coded Emirati speech it is ahead of that recogniser. It also yields a significantly lower CER on all six conditions of the further human-labelled sets. Even the frozen system's 128-best oracle stayed above Cohere's WER on the call-centre set. The adapted system's 16-best lists, by contrast, contain many such transcripts, so better rescoring is a natural next step.

\paragraph{What it cannot yet do} On long-form SaudiTalk speech, the adapted system still trails the dedicated recogniser in WER, although its CER there is lower. Our readers are bidirectional. Unlike Nemotron, neither system streams, even though the encoder's attention is chunk-wise, with no attention beyond the current 480 ms chunk.

\paragraph{Limitations}
Every configuration was trained once, so we have no estimate of seed variance. The adapted system's training and language-model data also differ from the frozen system's, which means its gains cannot be attributed to the encoder adapters alone. Its decoding weights were tuned at an earlier stage on validation calls whose machine labels favour CTC over joint search, the opposite of every human-labelled set; a validation set with human labels would be a better guide. We did not compare against other frozen encoders, such as the Whisper or w2v-BERT encoders, using the same reader, nor against the full Gemma 4 model used as a recogniser. Such controls would show how much of the result is specific to this encoder. The call-centre training labels are machine transcripts from two systems, one of which is the strongest baseline, so the reader is partly distilled from it. The call-centre references are mostly unedited machine drafts, as Section~\ref{sec:setup} explains, and the windows were selected to be hard. The blind call-centre sample was corrected by one annotator, the author, and is small. The only public customer-service set with original human references is a single call, so the adapted system's results there rest on one speaker. The adapted system also trains on voicebot calls with machine labels. Telephone conditions are simulated with G.711 $\mu$-law only, whereas real networks add other codecs, packet loss and echo. The SADA and Casablanca licences forbid commercial use, so the SADA-trained readers and language model are research artefacts only. Speed was measured on one GPU type and one CPU type, and only our own system received inference optimisation; Cohere and Nemotron were run with their standard implementations.

\section{Conclusion}
\label{sec:conclusion}
To our knowledge, this is the first Arabic speech-recognition study of the audio encoder shared by Gemma 4 and EmbeddingGemma 2, which we verified, for Gemma~4 E2B, to be identical in both releases except for the audio-embedding projection into the text model. Kept frozen and read by a 22.6M-parameter CTC model, the encoder supports Saudi and Gulf recognition whose CER is not significantly different from that of a 2B dedicated recogniser on five of six human-labelled conditions. It does so at several times the CPU speed and about a third of the GPU memory. With 3.1M adapter parameters, a hybrid reader and joint decoding, and still without changing a single pretrained encoder weight, the system is not significantly different from the 2B recogniser in WER on a public Saudi customer-service call and on clean Emirati speech. It is significantly better on telephone-coded Emirati speech and has a significantly lower CER on all six conditions of the further human-labelled sets. In WER, it still trails on long-form Saudi speech. It is also significantly better than a 0.6B recogniser fine-tuned on SADA in WER on all six of those conditions. An optimised joint decoder runs it at about \R{fast.p4.rtfx} times real time on one GPU. On a blind, human-corrected public call-centre sample it has the lowest WER of the self-hostable systems among the \R{api.n_systems} tested, while three commercial APIs score better, significantly so only GPT-Transcribe. We further show that references built by correcting machine drafts can inflate the apparent quality of the drafting system. We recommend that such references be reported by origin.

\section*{Ethics and Data Statement}
The in-house call recordings were processed for this research and are not released; no transcript or audio excerpt is reproduced in
this paper, and all results are aggregate scores. The recordings were used under an agreement with the data owner for research. SADA (CC~BY-NC-SA), Casablanca (CC~BY-NC-ND), MASC (CC~BY~4.0), SaudiTalk and the two public call-centre samples were used for research
only, under their published terms; no model trained on non-commercial data is distributed for commercial use.

\section*{Code and Data Availability}
Code for both systems, including the fast joint decoder, and every result on public test sets are available at
\url{https://github.com/AmrKhalil9/frozen-embeddinggemma2-arabic-asr}, together with the sourced list prices. The in-house call
recordings, their transcripts, the results on them and the trained weights are not released.

\section*{Funding}
This work received no external funding.

\section*{Competing Interests}
The author declares no competing interests.

\section*{Use of AI Assistance}
During this work the author used Claude (Anthropic) to help write code and to draft and edit the manuscript, and used other AI
assistants to review the experimental plan. The author reviewed and verified all content and takes full responsibility for it. Every
number in the tables and text is generated by a script directly from the released result files.

\bibliographystyle{unsrtnat}
\bibliography{refs_verified,refs_extra,refs_extra2,refs_extra3,refs_apis}
\end{document}